\documentclass{llncs}
\usepackage{makeidx}  % allows for indexgeneration
\usepackage{times}
\usepackage{graphicx}
\usepackage{latexsym}
\usepackage{mathnotation}

\usepackage{graphicx,amssymb}
\usepackage{latexsym}
\usepackage[ruled,vlined,linesnumbered]{algorithm2e}
\usepackage{amsmath}
\usepackage[all]{xy}
\usepackage{multirow}
\usepackage{url}
\usepackage{color}
\usepackage[draft]{fixme}
\fxsetup{
nomargin,inline,index,
theme=color
}

\newcommand{\tax}{\mathit{ax}}

\newcommand{\mo}{\mathcal{O}}
\newcommand{\mt}{\mathcal{T}}
\newcommand{\ma}{\mathcal{A}}

\newcommand{\md}{\mathcal{D}}
\newcommand{\mb}{\mathcal{B}}
\newcommand{\Te}{P}
\newcommand{\Tne}{N}
\newcommand{\te}{p}
\newcommand{\tne}{n}
\newcommand{\qry}{Q}

\newcommand{\mD}{{\bf{D}}}

\begin{document}

\title{Direct computation of diagnoses for ontology debugging\thanks{The research project is funded by Austrian Science Fund (Project V-Know, contract 19996).}}

\author{Kostyantyn Shchekotykhin \and Philipp Fleiss \and Patrick Rodler \and Gerhard Friedrich}
\authorrunning{Kostyantyn Shchekotykhin et al.} % abbreviated author list (for running head)
\tocauthor{Kostyantyn Shchekotykhin, Philipp Fleiss, Patrick Rodler, Gerhard Friedrich}
\institute{Alpen-Adria Universit\"at, Klagenfurt, 9020 Austria \\ \email{firstname.lastname@aau.at}}

\maketitle
\bibliographystyle{splncs03}

\begin{abstract}
Modern ontology debugging methods allow efficient identification and localization of faulty axioms defined by a user while developing an ontology. The ontology development process in this case is characterized by rather frequent and regular calls to a reasoner resulting in an early user awareness of modeling errors. In such a scenario an ontology usually includes only a small number of conflict sets, i.e. sets of axioms preserving the faults. This property allows efficient use of standard model-based diagnosis techniques based on the application of hitting set algorithms to a number of given conflict sets. However, in many use cases such as ontology alignment the ontologies might include many more conflict sets than in usual ontology development settings, thus making precomputation of conflict sets and consequently ontology diagnosis infeasible. In this paper we suggest a debugging approach based on a direct computation of diagnoses that omits calculation of conflict sets. Embedded in an ontology debugger, the proposed algorithm is able to identify diagnoses for an ontology which includes a large number of faults and for which application of standard diagnosis methods fails. The evaluation results show that the approach is practicable and is able to identify a fault in adequate time.  
\end{abstract}

\section{Introduction}
Ontology development and maintenance relies on an ability of users to express their knowledge in form of logical axioms. However, the knowledge acquisition process might be problematic since a user can make a mistake in an axiom being modified or a correctly specified axiom can trigger a hidden bug in an ontology. These bugs might be of different nature and are caused by violation of  \emph{requirements} such as consistency of an ontology, satisfiability of classes, presence or absence of some  entailments. %The complexity of the modification process grows proportional with the size of an ontology and the number of experts working on it. 
In such scenarios as ontology matching the complexity of faults might be very high because multiple disagreements between ontological definitions and/or modeling problems  an be triggered by aliments at once. 

Ontology debugging tools~\cite{Kalyanpur.Just.ISWC07,friedrich2005gdm,Horridge2008} can simplify the development process by allowing their users specification of requirements to the intended (target) ontology. If some of the requirements are broken, i.e.\ an ontology $\mo$ is \emph{faulty}, the debugging tool can compute a set of axioms $\md \subseteq \mo$ called \emph{diagnosis}. An expert should remove or modify at least all axioms of a diagnosis in order to be able to formulate the \emph{target ontology} $\mo_t$. Nevertheless, in real-world scenarios debugging tools can return a set of alternative diagnoses $\mD$, since it is quite hard for a user to specify such set of requirements that allows formulation of the target ontology $\mo_t$ only. Consequently, the user has to differentiate between multiple diagnoses $\md \in \mD$ in order to find the \emph{target diagnosis} $\md_t$, which application allows formulation of the intended ontology.
Diagnosis discrimination methods \cite{Kalyanpur2006,Shchekotykhin2012} allow their users to reduce the number of diagnoses to be considered. The first approach presented in~\cite{Kalyanpur2006} uses a number of heuristics that rank diagnoses depending on the structure of axioms $\tax_i \in \md$, usage in test cases, provenance information, etc. Only diagnoses with the highest ranks are returned to the user. 
%The main problem of this approach is that a user should provide enough data for the heuristics prior to debugging session. Moreover, there might be situations when suggested heuristics are not able to capture the target diagnosis and another heuristic is required. To overcome these issues 
A more sophisticated approach suggested in~\cite{Shchekotykhin2012} identifies the target diagnosis by asking an oracle, like an expert or information extraction system, a sequence of questions: whether some axiom is entailed by the target ontology or not. Given an answer the algorithm removes all diagnoses that are inconsistent with it. Furthermore, the query is used to create an additional test case, which allows the search algorithm to prune the search space and reduce the number of diagnoses to be computed. Moreover, in order to speed up the computations the method approximates the set of all diagnoses with a set of $n$ leading diagnoses, i.e.\ the $n$ best diagnoses with respect to a given measure.

All the approaches listed above follow the standard model-based diagnosis approach~\cite{Reiter87} and compute diagnoses using minimal conflict sets $CS$, i.e.\ irreducible sets of axioms $\tax_i \in \mo$ that preserve violation of at least one requirement. 
%Consequently, removal of at least one axiom from $CS$ results in a set $CS'$ that fulfills all requirements. 
The computation of the conflict sets can be done within a polynomial number of calls to the reasoner, e.g. by \textsc{QuickXPlain} algorithm~\cite{junker04}. To identify a diagnosis of cardinality $|\md|=m$ the hitting set algorithm suggested in~\cite{Reiter87} requires computation of $m$ minimal conflict sets. In the use cases when an ontology is generated by an ontology learning or matching system the number of minimal conflict sets $m$ can be large, thus making the ontology debugging practically infeasible. 

In this paper we present two algorithms \textsc{Inv-HS-Tree} and \textsc{Inv-QuickXPlain}, which inverse the standard model-based approach to ontology debugging and compute diagnoses directly, rather than by means of minimal conflict sets. Thus, given some predefined number of leading diagnoses $n$ the breadth-first search algorithm \textsc{Inv-HS-Tree} executes the direct diagnosis algorithm \textsc{Inv-QuickXPlain} exactly $n$ times. This property allows the new search approach to perform well when applied to ontologies with large number of conflicts. 
The evaluation shows that the direct computation of diagnoses allows to apply ontology debugging in the scenarios  suggesting diagnosis of generated ontologies. The system based on \textsc{Inv-QuickXPlain} is able to compute diagnoses and identify the target one in the cases when common model-based diagnosis techniques fail. Moreover, the suggested algorithms are able to maintain a comparable or a slightly better performance in the cases that can be analyzed by both debugging strategies.
%
%Moreover, in some cases application of the direct diagnosis computation to diagnosis discrimination leads to improvements in both overall debugging time and number of queries. 

The remainder of the paper is organized as follows: Section~\ref{sec:diag} gives a brief introduction to the main notions of ontology debugging. The details of the suggested algorithms and their application are presented in Section~\ref{sec:details}. In Section~\ref{sec:eval} we provide evaluation results and conclude in Section~\ref{sec:conc}.

\section{Ontology debugging} \label{sec:diag}

Let us exemplify the ontology debugging process by the following use case:
\begin{example} \label{ex:simple} Consider an ontology $\mo$ with the terminology $\mt$:
\begin{center}
\begin{tabular}{p{4cm}p{4cm}l}
$\tax_1 : A \sqsubseteq B$ & $\tax_2 : B \sqsubseteq E$ &  $\tax_3 : B \sqsubseteq D \sqcap \lnot \exists s.C$\\
$\tax_4 : C \sqsubseteq \lnot(D \sqcup E)$ & $\tax_5 : D \sqsubseteq \lnot B$& \\
\end{tabular}
\end{center}
and assertions $\ma :\{A(w), A(v), s(v,w)\}$. Because of the axioms $\tax_3$ and $\tax_5$ the given terminology is incoherent and it includes two unsatisfiable classes $A$ and $B$. Moreover, the assertions $A(w)$ and $A(v)$ make the ontology inconsistent. 

Assume that the user is sure that the assertional axioms are correct, i.e. this part of the ontology is included to the \emph{background knowledge} $\mb$ of the debugging system and, thus, cannot be considered as faulty in the debugging process. In this case, the debugger can identify the only irreducible set of axioms $CS': \tuple{\tax_3,\tax_5}$ -- minimal conflict set -- preserving both inconsistency and incoherency of $\mo$. Modification of at least one axiom of one of the minimal (irreducible) diagnoses $\md'_1: \diag{\tax_3}$ or $\md'_2: \diag{\tax_5}$ is required in order to restore both consistency and coherency of the ontology. 

In some debugging systems, e.g.~\cite{Shchekotykhin2012}, the user can also provide \emph{positive} $\Te$ and \emph{negative} $\Tne$ test cases, where each test case is a set of axioms that should be entailed (positive) and not entailed (negative) by the ontology resulting in application of the debugger. If in the example the user specifies 
\begin{equation*}
\Te=\setof{\setof{B(w)}} \text{ and } \Tne=\setof{\setof{\lnot C(w)}}
\end{equation*} 
then the debugger returns another set of minimal conflicts sets:
\begin{equation*}
CS_1: \tuple{\tax_1,\tax_3} \quad  CS_2: \tuple{\tax_2,\tax_4} \quad CS_3: \tuple{\tax_3,\tax_5}\quad CS_4: \tuple{\tax_3, \tax_4} 
\end{equation*}
and diagnoses:
\begin{equation*}
\md_1: \diag{\tax_3,\tax_4} \quad \md_2: \diag{\tax_2,\tax_3} \quad \md_3: \diag{\tax_1,\tax_4,\tax_5}
\end{equation*}
The reason is that both ontologies $\mo'_1 = \mo \setminus \md'_1$ and $\mo'_2 = \mo \setminus \md'_2$ resulting in application of the diagnoses $\md'_1$ and $\md'_2$ do not fulfill the test cases. For instance, $\mo'_2$ is invalid since the axioms $\setof{\tax_1,\tax_3} \subset \mo'_2$ entail $\lnot C(w)$, which must not be entailed. 
\end{example}

\begin{definition}[Target ontology]\label{def:target}
The target ontology $\mo_t$ %resulting in the application of an ontology debugger 
%to a diagnosis problem instance $\tuple{\mo,\mb,\Te,\Tne}$ 
is a set of axioms that is characterized by a background knowledge $\mb$, sets of positive $\Te$ and negative $\Tne$ test cases. The target ontology $\mo_t$ should fulfill the following necessary requirements\footnote{In the following we assume that the user intends to formulate only one ontology. In the paper we refer to the intended ontology as the target ontology.}:
\begin{itemize}
	\item $\mo_t \cup \mb$ must be consistent and, optionally, coherent
    \item $\mo_t \cup \mb \models\te \qquad \forall \te \in \Te$
    \item $\mo_t \cup \mb \not\models\tne \qquad \forall \tne \in \Tne$
\end{itemize}
The ontology $\mo$ is \emph{faulty} with respect to a predefined $\mb, \Te$ and $\Tne$ iff $\mo$ does not fulfill the necessary requirements. 
\end{definition}

\begin{definition}[Diagnosis problem instance]\label{def:diagproblem}
$\tuple{\mo,\mb,\Te,\Tne}$ is a \emph{diagnosis problem instance}, where $\mo$ is a faulty ontology, $\mb$ is a background theory, $\Te$ is a set of test cases that must be entailed by the target ontology $\mo_t$ and $\Tne$ is a set of test cases that must not be entailed by $\mo_t$. The instance is \emph{diagnosable} if $\mb\cup \bigcup_{\te \in \Te} \te$ is consistent (coherent) and $\mb\cup \bigcup_{\te \in \Te} \te \not\models \tne$ for each $\tne\in\Tne$.
\end{definition}

The ontology debugging approaches~\cite{Horridge2008,Kalyanpur.Just.ISWC07,Shchekotykhin2012} can be applied to any knowledge representation language for which there is a sound and complete procedure for deciding whether an ontology entails an axiom or not. Moreover, the entailment relation $\models$ \emph{must} be extensive, monotone and idempotent.

Another important aspect of the ontology debugging systems comes from the model-based diagnosis techniques~\cite{Reiter87,dekleer1987} that they are based on. The model-based diagnosis theory considers the modification operation as a sequence of add/delete operations and focuses only on deletion. That is, if an ontology includes faulty axiom then the simplest way to remove the fault is to remove the axiom. However, removing an axiom might be a too coarse modification, since the ontology can lose some of the entailments that must be preserved. Therefore, the model-based diagnosis takes also into account axioms (ontology extension) $EX$, which are added by the user to the ontology after removing all axioms of a diagnosis. Usually the set of axioms $EX$ is either formulated by the user or generated by a learning system~\cite{Lehmann2010}. If $EX$ is not empty then all axioms $\tax_i \in EX$  are added to the set of positive test cases $\Te$, since each axiom $\tax_i$ must be entailed by the intended ontology $\mo_t$.

\begin{definition}[Diagnosis]\label{def:diag}
For a diagnosis problem instance $\tuple{\mo,\mb,\Te,\Tne}$ a subset of the ontology axioms $\md\subset\mo$ is a \emph{diagnosis} iff 
%there is an extension $EX$ such that $\mo_t = (\mo\setminus\md)\cup EX$ 
$\mo_t = (\mo\setminus\md)$ fulfills the requirements of Definition~\ref{def:target}.
\end{definition}

Due to computational complexity of the diagnosis problem, in practice the set of all diagnoses is approximated by the set of minimal diagnoses. For a diagnosis problem instance $\tuple{\mo,\mb,\Te,\Tne}$ a diagnosis $\md$ is minimal iff there is no diagnosis $\md'$ of the same instance such that $\md'\subset\md$.

The computation of minimal diagnoses in model-based approaches is done by means of conflict sets, which are used to constrain the search space.
\begin{definition}[Conflict set]\label{def:conf}
For a problem instance $\tuple{\mo,\mb,\Te,\Tne}$ a set of axiom $CS\subseteq\mo$ is a \emph{conflict set} iff one of the conditions holds:
\begin{itemize}
	\item $CS\cup\mb\cup\bigcup_{\te \in \Te}$ is inconsistent (incoherent) or 
    \item $\exists\tne\in\Tne$ such that $CS\cup\mb\cup\bigcup_{\te \in \Te}\models\tne$
\end{itemize} 
\end{definition}
Just as for diagnoses, computation of conflict sets is reduced to computation of minimal conflict sets. A conflict set $CS$ is \emph{minimal} iff there is no conflict set $CS'$ such that $CS'\subset CS$.

\paragraph{Computation of minimal conflict sets.} In practice, the diagnosis systems use two types of strategies for computation of conflict sets, namely, brute-force~\cite{junker01,Kalyanpur.Just.ISWC07} and divide-and-conquer~\cite{junker04}. The first strategy can be split into acquisition and minimization stages. During the acquisition stage the algorithm adds axioms of an ontology $\mo\setminus\mb$ to a buffer while a set of axioms in the buffer is not a conflict set. As soon as at least one conflict set is added to the buffer, the algorithm switches to the minimization stage. In this stage axioms are removed from the buffer such that the set of axioms in the buffer remains a conflict set after each deletion. The algorithm outputs a minimal conflict set or 'no conflicts'. In the worst case a brute force algorithm requires $O(m)$ calls to the reasoner, where $m$ is the number of axioms in a faulty ontology.
The algorithm implementing divide-and-conquer strategy starts with a buffer containing all axioms of an ontology, i.e. the conflict set is in the buffer, and splits it into smaller and simpler sub-problems. The algorithm continues splitting until it identifies a sequence of sub-problems including only one axiom such that a set including all these axioms is a minimal conflict set. The divide-and-conquer algorithm requires in the worst case $O(k\log(\frac{m}{k}))$ calls to the reasoner, where $k$ is the cardinality of a returned conflict set. Taking into account that in practice $k\ll m$, the divide-and-conquer strategy is preferred to the brute-force.

\paragraph{Identification of minimal diagnoses.} The computation of minimal diagnoses in modern ontology debugging systems is implemented using the Reiter's Hitting Set \textsc{HS-Tree} algorithm~\cite{Reiter87,greiner1989correction}. The algorithm constructs a directed  tree from root to the leaves, where each node $nd$ is labeled either with a minimal conflict set $CS(nd)$ or $\checkmark$ (consistent) or $\times$ (pruned). The latter two labels indicate that the node is closed. Each edge outgoing from the open node $nd$ is labeled with an element $s \in CS(nd)$. $HS(nd)$ is a set of edge labels on the path from the root to the node $nd$.
Initially the algorithm creates an empty root node and adds it to the \emph{queue}, thus, implementing a breadth-first search strategy. Until the queue is empty, the algorithm retrieves the first node $nd$ from the queue and labels it with either:
\begin{enumerate}
    \item $\times$ if there is a node $nd'$, labeled with either $\checkmark$ or $\times$, such that $H(nd')\subseteq H(nd)$ (pruning non-minimal paths), or
    \item $CS(nd')$ if a node $nd'$ exists such that its label $CS(nd') \cap H(nd) = \emptyset$ (reuse), or
    \item $CS$ if $CS$ is a minimal conflict set computed for the diagnosis problem instance $\tuple{\mo \setminus H(nd),\mb,\Te,\Tne}$ by one of the algorithms mentioned above (compute), or
    \item $\checkmark~~$ (consistent).
\end{enumerate}
The leaf nodes of a complete tree are either pruned ($\times$) or consistent ($\checkmark$) nodes. The set of labels $H(nd)$ of each consistent node $nd$ corresponds to a minimal diagnosis. The  minimality of the diagnoses is guaranteed due to the minimality of conflict sets, pruning rule and breadth-first search strategy. Moreover, because of the latter the minimal diagnoses are generated in order of increasing cardinality.

\paragraph{Diagnoses discrimination.} In many real-world scenarios an ontology debugger can return a large number of diagnoses, thus, placing the burden of diagnosis discrimination on the user. Without an adequate tool support the user is often unable to understand the difference between the minimal diagnoses and to select an appropriate one. The diagnosis discrimination method suggested in~\cite{Shchekotykhin2012} uses the fact that different ontologies, e.g.\ $\mo_1=\mo\setminus\md_1$ and $\mo_2=\mo\setminus\md_2$, resulting in the application of different diagnoses, entail different sets of axioms. Consequently, there exists a set of axioms $\qry$ such that $\mo_1\models\qry$ and $\mo_2\not\models\qry$. 
%For instance, in the trivial case the set $\qry \subseteq \md_2\setminus\md_1$ can fulfill the condition if $\mo_1\models\qry$ since $\qry \subseteq \mo_1$ and  $\mo_2\not\models\qry$ since $\qry \not\subseteq \mo_2$  . 
If such a set of axioms $\qry$ exists, it can be used as a query to some oracle such as the user or an information extraction system. If the oracle answers $yes$ then the target ontology $\mo_t$ should entail $\qry$ and, hence, $\qry$ should be added to the set of positive test cases $\Te\cup\setof{\qry}$. Given the answer $no$ the set of axioms is added to the negative test cases $\Tne\cup\setof{\qry}$ to ensure that the target ontology does not entail $\qry$. Thus, in the first case the set of axioms $\md_2$ can be removed from the set of diagnoses $\mD$ because $\md_2$ is not a diagnosis of the updated diagnosis problem instance $\tuple{\mo,\mb,\Te\cup\setof{\qry},\Tne}$ according to Definition~\ref{def:diag}. Similarly, in the second case the set of axioms $\md_1$ is not a diagnosis of $\tuple{\mo,\mb,\Te,\Tne\cup\setof{\qry}}$. 

However, many different queries might exist for the set of diagnoses $|\mD| > 2$. In the extreme case there are $2^n-2$ possible queries for a set of diagnoses including $n$ elements. To select the best query the authors in~\cite{Shchekotykhin2012} suggest two measures: \textsc{split-in-half} and \textsc{entropy}. The first measure is a greedy approach preferring the queries which allow to remove a half of the minimal diagnoses from $\mD$, given an answer of an oracle. The second is an information-theoretic measure, which estimates the information gain for both outcomes of each query and returns the one that maximizes the information gain. The \emph{prior fault probabilities} required for \textsc{Entropy} measure can be obtained from statistics of previous diagnosis sessions. For instance, if the user has problems with understanding of restrictions then the diagnosis logs will contain more repairs of axioms including restrictions. Consequently, the prior fault probabilities of axioms including restrictions should be higher. Given the fault probabilities of axioms, one can calculate prior fault probabilities of minimal diagnoses including these axioms as well as evaluate \textsc{Entropy} (see~\cite{Shchekotykhin2012} for more details). 

A general algorithm of the interactive ontology diagnosis process can be described as follows:
\begin{enumerate}
	\item Generate a set of diagnoses $\mD$ including at most $n$ diagnoses.
    \item Compute a set of queries and select the best one according to some predefined measure.
    \item Ask the oracle and, depending on the answer, add the query either to $\Te$ or to $\Tne$.
    \item Update the set of diagnoses $\mD$ and remove the ones that do not comply with the newly acquired test case, according to the Definition~\ref{def:diag}.    
    \item Update the tree and repeat from Step 1 if the queue contains open nodes.
    \item Return the set of diagnoses $\mD$.
\end{enumerate}
The resulting set of diagnoses $\mD$ includes only diagnoses that are not differentiable in terms of their entailments, but have some syntactical differences. The preferred diagnosis in this case should be selected by the user using some text differencing and comparison tool. 

Note, that a similar idea can be found in~\cite{Nikitina2011} where authors use queries to an oracle to revise an ontology. Given a consistent and coherent ontology $\mo$ the system partitions it into two ontologies $\mo^{\models}$ and $\mo^{\not\models}$ containing required and incorrect  consequences correspondingly. The system can deal with inconsistent/incoherent ontologies if a union of all minimal conflict sets is put to the initial set of incorrect consequences $\mo^{\not\models}_0$. Computation of the set $\mo^{\not\models}_0$ requires application of an ontology debugger and is not addressed in~\cite{Nikitina2011}.

\paragraph{Example 1, continued.}  Assume that the user mistakenly negated only the restriction in $\tax_3$ instead of the whole description $B \sqsubseteq \lnot (D \sqcap \exists s.C)$. Moreover, in $\tax_4$ a disjunction was placed instead of conjunction because of a typo, i.e.\ $C \sqsubseteq \lnot(D \sqcap E)$.  

\begin{figure}[bt]
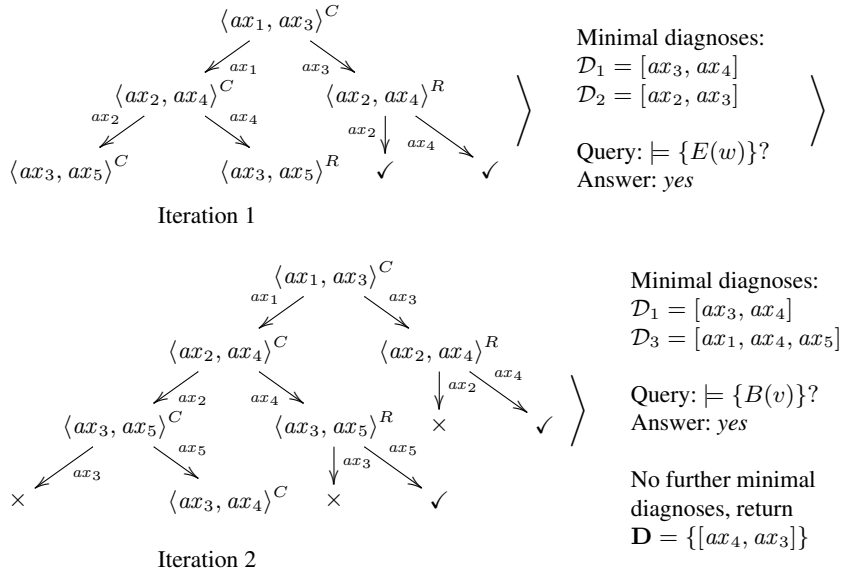

\begin{minipage}[c]{0.45\textwidth} 
\xygraph{
!{<0cm,0cm>;<1.4cm,0cm>:<0cm,1cm>::}
!{(4,0) }*+{\checkmark}="d2"
!{(3,0) }*+{\checkmark}="d1"
!{(1,1) }*+{\left<\tax_2, \tax_4\right>^C}="c2c"
!{(3,1) }*+{\left<\tax_2, \tax_4\right>^R}="c2r"
!{(0,0) }*+{\left<\tax_3, \tax_5\right>^C}="c3c"
!{(2,0) }*+{\left<\tax_3, \tax_5\right>^R}="c3r"
!{(2,2)}*+{\left<\tax_1, \tax_3\right>^C}="c1c"
"c2c":"c3r"^{\tax_4}
"c2c":"c3c"_{\tax_2}
"c1c":"c2c"^{\tax_1}
"c1c":"c2r"_{\tax_3}
"c2r":"d1"_{\tax_2}
"c2r":"d2"_{\tax_4}
}
\vspace{5pt}
\begin{center}
Iteration 1
\end{center}

\end{minipage}
\begin{minipage}[c]{20pt}
$\Bigg>$ 
\end{minipage}
\begin{minipage}[c]{0.25\textwidth}
\begin{tabular}{l}                                          
    Minimal diagnoses: \\
                $\md_1=[\tax_3, \tax_4]$ \\
                $\md_2=[\tax_2, \tax_3]$ \\
         \\
                Query: $\models \setof{E(w)}?$ \\
                Answer: \emph{yes}
		\end{tabular}
\end{minipage}
\begin{minipage}[c]{10pt}
$\Bigg>$ 
\end{minipage} 

\vspace{10pt}
\begin{minipage}[c]{0.45\textwidth}
\centering
\xygraph{
!{<0cm,0cm>;<1.4cm,0cm>:<0cm,1cm>::}
!{(4,0)}*+{\checkmark}="d2"
!{(4,1)}*+{\times}="o4"
!{(0,0)}*+{\times}="o2"
!{(5,1)}*+{\checkmark}="d1"
!{(3,0)}*+{\times}="o1"
!{(3,1)}*+{\left<\tax_3, \tax_5\right>^R}="c3r"
!{(4,2)}*+{\left<\tax_2, \tax_4 \right>^R}="c2r"
!{(2,0)}*+{\left<\tax_3, \tax_4\right>^C}="c4"
!{(1,1)}*+{\left<\tax_3, \tax_5\right>^C}="c3"
!{(2,2)}*+{\left<\tax_2, \tax_4 \right>^C}="c2"
!{(3,3)}*+{\left<\tax_1, \tax_3\right>^C}="c1"
"c3r":"d2"^{\tax_5}
"c3r":"o1"^{\tax_3}
"c3":"c4"^{\tax_5}
"c3":"o2"^{\tax_3}
"c2r":"d1"^{\tax_4}
"c2r":"o4"^{\tax_2}
"c2":"c3r"_{\tax_4}
"c2":"c3"^{\tax_2}
"c1":"c2r"^{\tax_3}
"c1":"c2"_{\tax_1}
}
\vspace{10pt}
\begin{center}
Iteration 2
\end{center}
\end{minipage}
\begin{minipage}[c]{20pt}
$\Bigg>$ 
\end{minipage} 
\begin{minipage}[c]{0.25\textwidth}
		\begin{tabular}{l}  
                Minimal diagnoses: \\
                $\md_1=[\tax_3, \tax_4]$ \\
                $\md_3=[\tax_1, \tax_4, \tax_5]$ \\
                \\
                Query: $\models \setof{B(v)}?$ \\
                Answer: \emph{yes} \\
                \\
                No further minimal \\
                diagnoses, return \\
                $\mD = \setof{[\tax_4, \tax_3]}$
		\end{tabular}
\end{minipage}
\caption{Identification of the target diagnosis $[\tax_4,\tax_3]$ using diagnosis discrimination presented in \cite{Shchekotykhin2012} (HS-Tree and QuickXPlain computing conflicts on-demand). On each iteration two diagnoses ($n=2$) were computed  to identify a query. The answer was used to prune the search tree. All computed labels are denoted with $C$ and all reused with $R$.} \label{fig:hs:ex}
\end{figure}

The interactive diagnosis process, illustrated in Fig.~\ref{fig:hs:ex}, applies the three techniques described above to find the target diagnosis $\md_t=\diag{\tax_3,\tax_4}$. In the first iteration the system starts with the root node, which is labeled with $\tuple{\tax_1,\tax_3}$ -- the first conflict returned by \textsc{QuickXPlain}. Next, \textsc{HS-Tree} generates successor nodes and labels the edges leading to these nodes with corresponding axioms. The algorithm extends the search tree until two \emph{leading minimal diagnoses} are computed. Given the set of minimal diagnoses the diagnosis discrimination algorithm identifies a query using entailments of the two ontologies $\mo_1=\mo\setminus\md_1$ and $\mo_2=\mo\setminus\md_2$, which are deduced by the classification and realization services of a standard Description Logic reasoner. One of the entailments $E(w)$ can be used as a query, since $\mo_1\models E(w)$ and $\mo_2\not\models E(w)$. Given a positive answer of an oracle the algorithm updates the search tree and closes the node corresponding to the invalid minimal diagnosis $\md_2$. Since there are some open nodes, the algorithm continues and finds the next diagnosis $\md_3$. The two more nodes, expanded by the \textsc{HS-Tree} in the second iteration, are closed since both sets of labels on the paths to these nodes from the root are supersets of the closed paths $\setof{\tax_3,\tax_2}$ and $\setof{\tax_3,\tax_4}$. For the two minimal diagnoses $\md_1$ and $\md_3$ the diagnosis discrimination finds a query $\qry=\setof{B(v)}$, which is answered positively by the oracle. Consequently, the algorithm removes $\md_3$ and continues with the expansion of the last node labeled with $\tuple{\tax_3, \tax_4}$. Since the paths to the successors of this node are supersets of existing closed paths in the tree, the algorithm closes these nodes and  terminates. The diagnosis $\md_1$ suggesting modification of the axioms $\tax_4$ and $\tax_3$ is returned to the user.

The example shows that a modern ontology debugger can efficiently identify the target diagnosis. As it is demonstrated by different evaluation studies~\cite{Kalyanpur.Just.ISWC07,Shchekotykhin2012}, the debuggers work well in an ontology development and maintenance process in which users modify an ontology manually. In such process the users classify ontology regularly and, therefore, are able to identify the presence of faults early, i.e.\ a user introduces only a small number of modifications to the ontology between two calls to a reasoner. Therefore, faulty ontologies in this scenario can be characterized by a small number of minimal conflict sets that can generate a large number of possible diagnoses. For instance, Transportation ontology (see~\cite{Shchekotykhin2012})
includes only $9$ minimal conflict sets that generate $1782$ minimal diagnoses. In such case \textsc{HS-Tree} makes only $9$ calls to \textsc{QuickXPlain} and then reuses the identified minimal conflicts to label all other nodes. The number of calls to the reasoner, which is the main ``source of complexity'', is rather low and can be approximated by $9k\log{\frac{n}{k}}+ |Nodes|$, where $|Nodes|$ is the cardinality of the set containing all nodes of the search tree, $k$ is the maximum cardinality of all computed minimal conflict sets and $n$ is the number of axiom in the faulty ontology. The combination of ontology debugging with diagnosis discrimination allows to reduce the number of calls to the reasoner, because often acquired test cases invalidate not only diagnoses that are already computed by \textsc{HS-Tree}, but also those that are not. All these factors together with such techniques as module extraction~\cite{SattlerSZ09} make the application of ontology debuggers feasible in the described scenario.

However, in such applications as ontology matching or learning the number of minimal conflict sets can be much higher, because all axioms are generated at once. For instance, ontology alignments, identified by most of ontology matching systems in the last Ontology Alignment Evaluation Initiative (OAEI), are often incoherent and, in some cases, inconsistent~\cite{Ferrara2011}. The large number of minimal conflict set makes the application of ontology debugging problematic because of the large number of calls to the reasoner and the memory required by the breadth-first search algorithm. To overcome this problem we suggest a novel ontology debugging approach that computes diagnoses directly, i.e.\ without precomputation of minimal conflict sets.

%In the worst-case the number of minimal conflict sets might be exponential in the number of axioms~\cite{de2004fundamentals}. 

\section{Direct diagnosis of ontologies}\label{sec:details}

%Dual QuickXPlain
The main idea behind the approach is to start with the set $\md_0=\emptyset$ and extend it until such a subset of ontology axioms $\md \subseteq \mo$ is found that $\md$ is a minimal diagnosis with respect to the Definition~\ref{def:diag}. In the first step Algorithm~\ref{algo:qx} verifies the input data, i.e.\ if the input ontology $\mo$, background knowledge $\mb$, positive $\Te$ and negative $\Tne$ test cases together constitute a valid diagnosis problem instance $\tuple{\mo,\mb,\Te,\Tne}$ (Definition~\ref{def:diagproblem}). Thus it verifies: a) whether the background theory together with the positive and negative test cases is consistent; and b) if the ontology is faulty. In both cases \textsc{Inv-QuickXPlain} calls \textsc{verifyRequirements} function that implements Definition~\ref{def:diag} and tests if axioms in the set $\md$ are a minimal diagnosis or not. The test function requires a reasoner that implements consistency/coherency checking (\textsc{isConsistent}) and allows to decide whether a set of axioms is entailed by the ontology (\textsc{entails}). 

\textsc{findDiagnosis} is the main function of algorithm which takes six arguments as input. The values of the arguments $\mb$, $\mo$ and $\Tne$ remain constant during the recursion and are required only for verification of requirements.  Whereas values of $\md$, $\Delta$ and $\mo_\Delta$ are used to provide a set of axioms corresponding to the actual diagnosis and two diagnosis sub-problems on the next level of the recursion. The sub-problems are constructed during the execution \textsc{findDiagnosis} by splitting a given diagnosis problem with \textsc{split} function. In the most of the implementations \textsc{split} simply partitions the set of axioms into two sets of equal cardinality. The algorithm continues to divide diagnosis problems (\textsc{findDiagnoses} line 12) until it identifies that the set $\md$ is a diagnosis (line 7). In further iterations the algorithm  minimizes the diagnosis by splitting it into sub-problems of the form $\md = \md' \cup \mo_\Delta$, where $\mo_\Delta$ contains only one axiom. In the case when $\md$ is a diagnosis and $\md'$ is not, the algorithm decides that $\mo_\Delta$ is a subset of the sought minimal diagnosis. Just as the original algorithm, \textsc{Inv-QuickXPlain} always terminates and returns a minimal diagnosis for a given diagnosis problem instance.

\begin{algorithm}[bt]
\caption{\textsc{Inv-QuickXPlain$(\mo, \mb, \Te, \Tne)$}} \label{algo:qx}
\KwIn{$\mo$ set of faulty axioms, $\mb$ set of background axioms, $\Te$ set of positive test cases, $\Tne$ set of negative test cases}
\KwOut{a minimal diagnosis $\md$} 
\SetKwFunction{vr}{\normalfont \textsc{verifyRequirements}}
\SetKwFunction{generate}{\normalfont \textsc{findDiagnosis}}
\SetKwFunction{split}{\normalfont \textsc{split}}
\SetKwFunction{get}{\normalfont \textsc{getElements}}
\SetKwFunction{iscons}{\normalfont \textsc{isConsistent}}
\SetKwFunction{entails}{\normalfont \textsc{entails}}

\SetKwBlock{part}{function {\normalfont \textsc{findDiagnosis} ($\mb, \md, \Delta, \mo_\Delta, \mo, \Tne$)} returns {\normalfont a minimal diagnosis $\md$}}{end}

\SetKwBlock{ver}{function {\normalfont \textsc{verifyRequirements} ($\mb, \md, \mo, \Tne$)} returns {\normalfont \emph{true} or \emph{false}}}{end}

$\mo' \leftarrow \mo \setminus\mb$ \;
$\mb' \leftarrow \mb \cup \bigcup_{\te\in\Te} \te$ \;

\lIf{$\lnot \vr(\mb', \emptyset, \emptyset, \Tne)$}{
    \Return~'inconsistent requirements'}\;

\lIf{$\vr(\mb', \emptyset, \mo', \Tne)$}{
    \Return~'consistent'}\;
 
\Return $\generate(\mb', \emptyset, \mo', \mo', \Tne)$\; \vspace{7pt}

\part{
    \lIf {$\Delta \neq \emptyset \land \vr(\mb, \md, \mo, \Tne)$}{                
        \Return $\emptyset$\;
    }
    \lIf {$|\mo| = 1$}{                
        \Return $\mo$\;
    }     
    $k \leftarrow \split(\mo)$\;
    $\mo_1 \leftarrow \get(\mo_\Delta, 1, k)$\;
    $\mo_2 \leftarrow \get(\mo_\Delta, k - 1, |\mo_\Delta|)$\;
    
    %$\mb \leftarrow \mb\cup\mo_1$\;    
    $\md_2 \leftarrow \generate(\mb, \md\cup\mo_1, \mo_1, \mo_2, \mo, \Tne)$\;
    %$\mb \leftarrow (\mb\setminus\mo_1)\cup\md_2$\;       
    
    $\md_1 \leftarrow \generate(\mb, \md\cup\md_2, \md_2, \mo_1, \mo, \Tne)$\; 
    %$\mb \leftarrow$ $\mb\setminus\md_2$\;
    
    \Return $\md_1 \cup \md_2$\;
}

\ver{    
    $\mo' \leftarrow \mb \cup (\mo \setminus \md)$ \;
    $c \leftarrow \iscons(\mo')$ \;
    \lIf{$\lnot c$}{\Return false}\;
    \lForEach{$\tne \in \Tne$}{$c \leftarrow c \land \entails(\mo', \tne)$}\;
    \Return $c$ \;
}
\end{algorithm}

\begin{figure}[tb]
	\centering
\includegraphics[width=.9\textwidth]{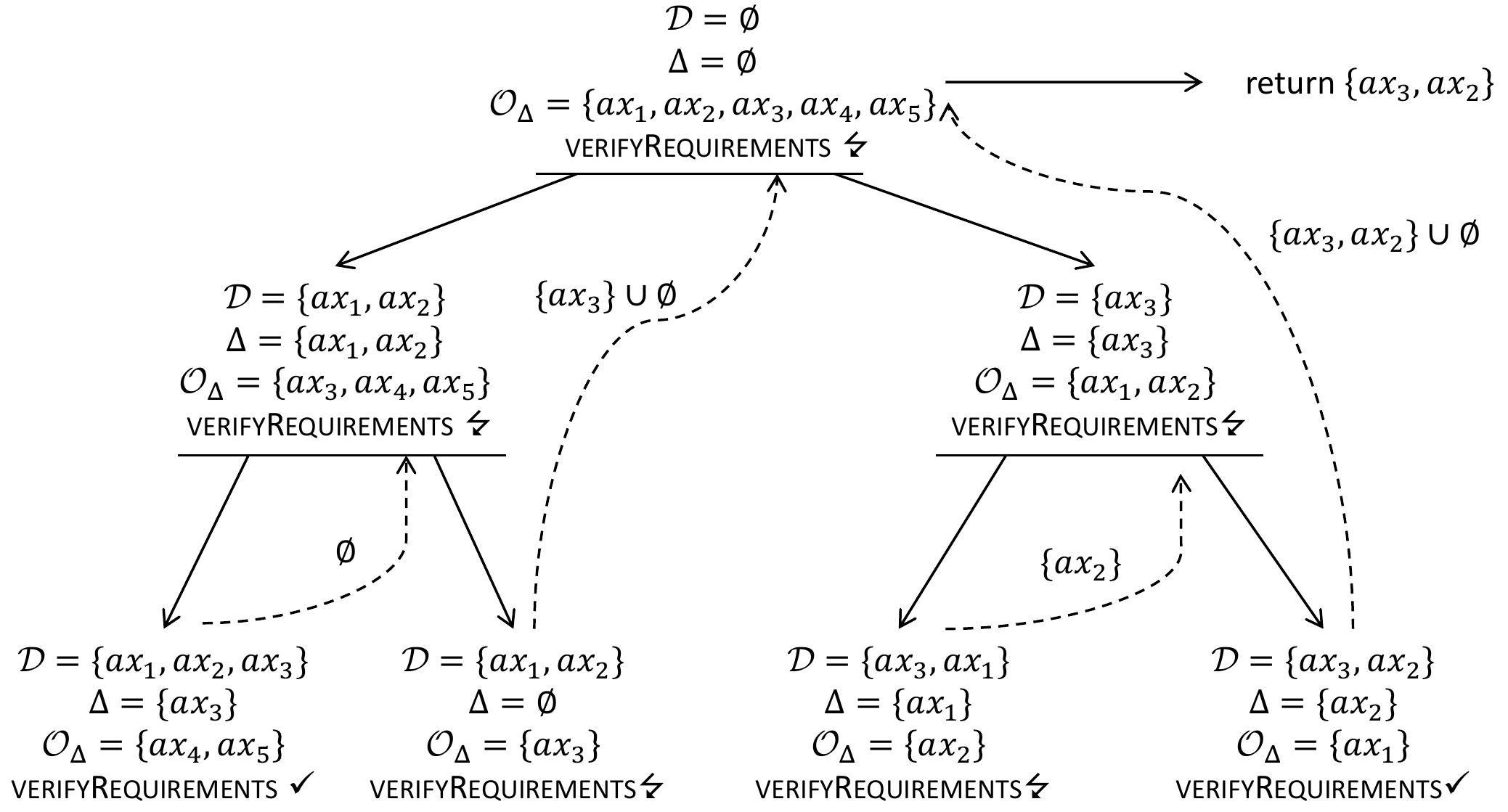}
\caption{Recursive calls of \textsc{Inv-HS-Tree} for the diagnosis problem instance in Example~\ref{ex:simple}. The background theory $\mb$, original ontology $\mo$ and the set of negative test cases $\Tne$ remain constant and therefore are omitted. Solid arrows show recursive calls (line 12 -- left and line 13 -- right) and dashed arrows indicate returns.}
\label{fig:qx}
%\vspace{-10pt}
\end{figure}

\paragraph{Example~\ref{ex:simple}, continued.} Let us look again at the ontology diagnosis example and show how a diagnosis is computed by \textsc{Inv-QuickXPlain} (see Fig.~\ref{fig:qx}). The algorithm starts with an empty diagnosis $\md=\emptyset$ and $\mo_\Delta$ containing all axioms of the problem. \textsc{verifyRequirements} returns $false$ since the $\mb \cup \mo \setminus \emptyset$ is inconsistent. Therefore, the algorithm splits $\mo_\Delta$ into $\setof{\tax_1,\tax_2}$ and $\setof{\tax_3,\tax_4,\tax_5}$ and passes the sub-problem to the next level of recursion. Since, the set $\md=\setof{\tax_1,\tax_2}$ is not a diagnosis, the ontology $\mb \cup (\mo \setminus \md)$ is inconsistent and the problem in $\mo_\Delta$ is split one more time. On the second level of recursion the set $\md$ is a diagnosis, although not minimal. The function \textsc{verifyRequirements} returns \emph{true} and the algorithm starts to analyze the found diagnosis. Therefore, it verifies whether the last extension of the set $\md$ is a subset of a minimal diagnosis. Since, the extension includes only one axiom $\tax_3$ an the extended set $\setof{\tax_1,\tax_2}$ is not a diagnosis, the algorithm concludes that $\tax_3$ is an element of the target diagnosis. The left-most branch of the recursion tree terminates and returns $\setof{\tax_3}$. This axiom is added to the set $\md$ and the algorithm starts investigating whether the two axioms $\setof{\tax_1,\tax_2}$ also belong to a minimal diagnosis. First, it tests the set $\setof{\tax_3,\tax_1}$, which is not a diagnosis, and on the next iteration it identifies the correct result $\setof{\tax_3,\tax_2}$.

\textsc{Inv-QuickXPlain} is a deterministic algorithm and returns the same minimal diagnosis if applied twice to a diagnosis problem instance. In order to obtain different diagnoses, the problem instance should be changed such that the \textsc{Inv-QuickXPlain} will identify the next diagnosis. Therefore, we suggest \textsc{Inv-HS-Tree}, which is a modification of the \textsc{HS-Tree} algorithm presented in Section~\ref{sec:diag}. The inverse algorithm labels each node $nd$ of the tree with a minimal diagnosis $\md(nd)$. The rules 1, 2 and 4 of the original algorithm remain the same in  \textsc{Inv-HS-Tree} and the rule 3 is modified as follows:
\begin{itemize}
	\item[3.] The open node $nd$ is labeled with $\md$ if $\md$ is a minimal diagnosis for the diagnosis problem instance $\tuple{\mo, \mb \cup H(nd),\Te,\Tne}$ computed by \textsc{Inv-QuickXPlain} (compute)
\end{itemize}
where $H(nd)$ is a set containing edge labels on the path from the root to $nd$. In this case elements of $H(nd)$ correspond to the axioms of minimal diagnoses that were used as labels of nodes on the path. Addition of an axiom $\tax_k$ of a minimal diagnosis $\md_i$ to the background theory forces \textsc{Inv-QuickXPlain} to search for a minimal diagnosis that suggests the modification of any other axiom, except $\tax_k$. That is, the diagnosis $\md_i$ will not be rediscovered by the direct diagnosis algorithm.

A modified update procedure is another important feature of \textsc{Inv-HS-Tree}. In the diagnosis discrimination settings the ontology debugger acquires new knowledge that can invalidate some of the diagnoses that are used as labels of the tree nodes. During the tree update \textsc{Inv-HS-Tree} searches for the nodes with invalid labels. Given such a node, the algorithm removes its label and places it to the list of open nodes.  Moreover, the algorithm removes all the nodes of a subtree originating from this node. After all nodes with invalid labels are cleaned-up, the algorithm attempts to reconstruct the tree by reusing the remaining valid minimal diagnoses (rule 2, \textsc{HS-Tree}). Such aggressive pruning of the tree is feasible since a) the tree never contains more than $n$ nodes that were computed with \textsc{Inv-QuickXPlain} and b) computation of a possible modification to the minimal diagnosis, that can restore its validness, requires invocation of \textsc{Inv-QuickXPlain} and, therefore, as hard as computation of a new diagnosis. Note also, that in a common diagnosis discrimination setting $n$ is often set to a small number, e.g.\ $10$, in order to achieve good responsiveness of the system.  Consequently, in this settings the size of the tree will be small. The latter is another advantage of the direct method as it requires much less memory in comparison to a debugger based on the breadth-first strategy. 

\begin{figure*}[tb]
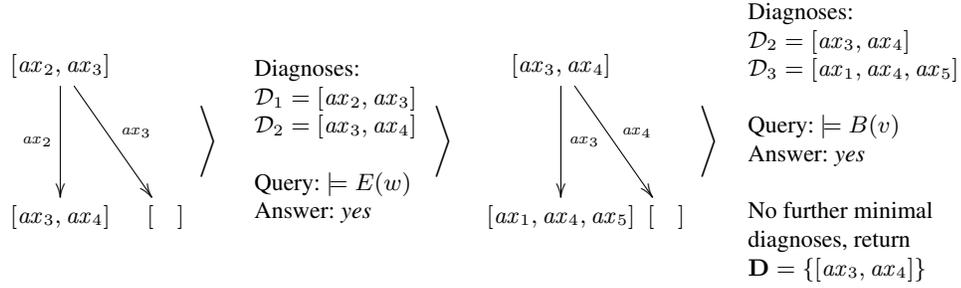

\centering
\begin{minipage}[c]{0.20\linewidth} 
\xygraph{
!{<0cm,0cm>;<1.4cm,0cm>:<0cm,1cm>::}
!{(1,0) }*+{[\quad]}="d1"
!{(0,0) }*+{[\tax_3, \tax_4]}="c3"
!{(0,2)}*+{[\tax_2, \tax_3]}="c1"
"c1":"d1"^{\tax_3}
"c1":"c3"_{\tax_2}
}
\end{minipage}
\begin{minipage}[c]{15pt}
$\Bigg>$ 
\end{minipage}
\begin{minipage}[c]{0.2\linewidth}
\centering
\begin{tabular}{l}                                          
      Diagnoses: \\
                $\md_1=[\tax_2, \tax_3]$ \\
                $\md_2=[\tax_3, \tax_4]$ \\
         \\
                Query: $\models E(w)$\\
                Answer: \emph{yes}
		\end{tabular}
\end{minipage}
\begin{minipage}[c]{15pt}
$\Bigg>$ 
\end{minipage} 
\begin{minipage}[c]{0.17\linewidth}
\xygraph{
!{<0cm,0cm>;<1.4cm,0cm>:<0cm,1cm>::}
!{(0,0) }*+{[\tax_1, \tax_4, \tax_5]}="d1"
!{(1,0) }*+{[\quad]}="d0"
!{(0,2) }*+{[\tax_3, \tax_4]}="c3"
"c3":"d0"^{\tax_4}
"c3":"d1"^{\tax_3}
}
\end{minipage}
\begin{minipage}[c]{15pt}
$\Bigg>$ 
\end{minipage}
\begin{minipage}[c]{0.17\linewidth}
\centering
\begin{tabular}{l}                                          
      Diagnoses: \\
                $\md_2=[\tax_3, \tax_4]$ \\
                $\md_3=[\tax_1, \tax_4, \tax_5]$ \\
         \\
                Query: $\models B(v)$\\
                Answer: \emph{yes} \\
         \\
         				No further minimal \\
                diagnoses, return \\
                $\mD = \setof{[\tax_3, \tax_4]}$
		\end{tabular}
\end{minipage}
\caption{Identification of the target diagnosis $[\tax_3,\tax_4]$ using direct diagnosis approach. 
} \label{fig:dual:ex}
%\vspace{-10pt}
\end{figure*}

\paragraph{Example~\ref{ex:simple}, continued.} Applied to the sample diagnosis problem instance the direct ontology debugger computes two minimal diagnoses $\diag{\tax_2, \tax_3}$ and $\diag{\tax_3, \tax_4}$ in the first iteration (see Fig.~\ref{fig:dual:ex}). For these diagnoses the discrimination method identifies the query $E(w)$, which is answered \emph{yes} by the oracle. The label of the root node in this case becomes invalid. Consequently, the algorithm removes the label of the root, deletes its subtree and places the root to the list of the open nodes. Next, according to the rule 2, the valid minimal diagnosis $\md_2$ is reused to label the root. Given a diagnosis problem instance $\tuple{\mo, \mb \cup \setof{\tax_3}, \Te, \Tne}$, \textsc{Inv-QuickXPlain} computes the last minimal diagnosis $\md_3$. Given the positive answer to the query $B(v)$ the algorithm labels both open nodes with $\checkmark$ and returns $\md_2$  as the result. Moreover, the labels on the edges of the tree correspond to the minimal conflicts $\tuple{\tax_3}$ and $\tuple{\tax_4}$ of the final diagnosis problem instance.

\section{Evaluation}\label{sec:eval}

We evaluated the direct ontology debugging technique using aligned ontologies generated in the framework of OAEI 2011~\cite{Ferrara2011}. These ontologies represent a real-world scenario in which a user generated ontology alignments by means of some (semi-)automatic tools. In such case the size of the minimal conflict sets and their configuration might be substantially different from the ones considered in the manual ontology development process, e.g.~\cite{Kalyanpur.Just.ISWC07,Shchekotykhin2012}. In the first experiment we demonstrate that \textsc{Inv-HS-Tree} is able to identify minimal diagnoses in the cases when \textsc{HS-Tree} fails. The second test shows that the direct diagnosis approach is scalable and can be applied to ontologies including thousands of axioms.

%%%%%%%%%%%%%%%%%%%%%%%%%%%%%%
% Ontology matching intro
The ontology matching problem can be formulated as follows: given two ontologies $\mo_i$ and $\mo_j$, the goal of the ontology matching system is to identify a set of alignments $M_{ij}$. Each element of this set is a tuple $\tuple{x_i,x_j,r,v}$, where $x_i \in Q(\mo_i)$, $x_j \in Q(\mo_j)$, $r$ is a semantic relation and $v$ is a confidence value. $Q(\mo)$ denotes a set of all matchable elements of an ontology $\mo$ such as classes or properties. The result of the ontology matching process is the aligned ontology $\mo_{ij} = \mo_i \cup M_{ij} \cup \mo_j$. In the ontologies, used in both experiments, only classes and properties were considered as matchable elements and the set of relations was limited to $r \in \setof{\sqsubseteq, \equiv, \sqsupseteq}$. 

%%%%%%%%%%%%%%%%%%%%%%%%%%%%%%

\begin{figure}[tb]
	\centering
		\includegraphics[width=.8\textwidth]{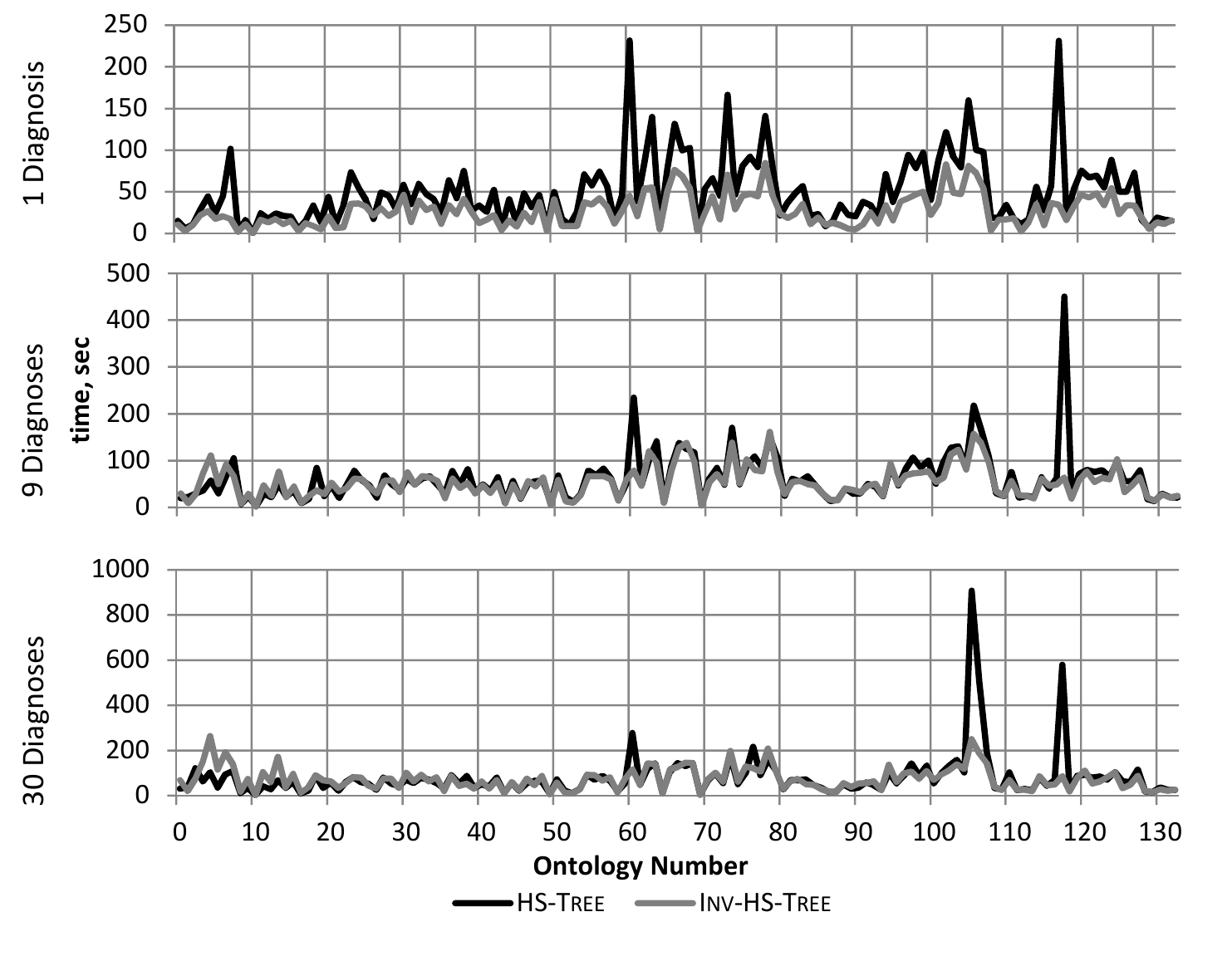}
        \caption{Time required to compute 1, 9  and 30 diagnoses by \textsc{HS-Tree} and \textsc{Inv-HS-Tree} for the Conference problem.}
        \label{fig:solvable}
\end{figure}

%%%%%%%%%%%%%%%%%%
In the first experiment we applied the debugging technique to the set of aligned ontologies resulting from ``Conference'' set of  problems, which is characterized by lower precision and recall of the applied systems (the best F-measure 0.65) in comparison, for instance, to the ``Anatomy'' problem (average F-measure 0.8)\footnote{see http://oaei.ontologymatching.org/2011.5/results/index.html for preliminary results of the evaluation based on reference alignments}.
The Conference test suite\footnote{All ontologies used in the evaluation can be downloaded from http://code.google.com/p/rmbd/wiki/DirectDiagnosis The webpage contains also tables presenting detailed results of the experiment presented in Fig.~\ref{fig:solvable}.} includes 286 ontology alignments generated by the 14 ontology matching systems. We tested all the ontologies of the suite and found that: a) 140 ontologies are consistent and coherent; b) 122 ontologies are incoherent; c) 26 ontologies are inconsistent; and in 8 cases HermiT~\cite{Motik2009} was unable to finish the classification in two hours\footnote{The tests were executed on a Core-i7 (3930K) 3.2Ghz, 32GB RAM and with Ubuntu Server 11.04, Java 6 and HetmiT 1.3.6 installed.}. The results show that only two systems \texttt{CODI} and \texttt{MaasMtch} out of 14 were able to generate consistent and coherent alignments. This observation confirms the importance of high-performance ontology debugging methods.

The 146 ontologies of the cases b) and c) were analyzed with both \textsc{HS-Tree} and \textsc{Inv-HS-Tree}. For each of the ontologies the system computed 1, 9 and 30 leading minimal diagnoses. The results of the experiment presented in Fig.\ref{fig:solvable} show that for 133 ontologies out of 146 both approaches were able to compute the required amount of diagnoses. In the experiment where only 1 diagnosis was requested, the direct approach outperforms the \textsc{HS-Tree} as it was expected. In the next two experiments the time difference between the approaches decreases. However, the direct approach was able to avoid a rapid increase of computation time for very hard cases. 

\begin{table}[tb]
\begin{center}
\begin{tabular}{|l|l|l|r|r|r|}
\hline
\textbf{Matcher} & \textbf{Ontology 1} & \textbf{Ontology 2} & \multicolumn{3}{c|}{\textbf{Time (ms) for }} \\ \cline{4-6}
& & & \textbf{1 Diag} & \textbf{9 Diags} & \textbf{30 Diags} \\
\hline
ldoa	& cmt	& edas	& 2540	& 6919	& 15470 \\
csa	& conference	& edas	& 3868	& 14637	& 39741 \\
ldoa	& ekaw	& iasted	& 10822	& 71820	& 229728 \\
mappso	& edas	& iasted	& 11824	& 89707	& 293746 \\
csa	& edas	& iasted	& 15439	& 134049	& 377361 \\
csa & conference & ekaw & 11257 & 31010 & 62823 \\
ldoa & cmt & ekaw & 5602 & 19730 & 42284 \\
ldoa & conference & confof & 8291 & 23576 & 48062 \\
ldoa & conference & ekaw & 7926 & 27324 & 56988 \\
mappso & conference & ekaw & 11394 & 33763 & 70469 \\
mappso & confof & ekaw & 9422 & 25921 & 55667 \\
optima & conference & ekaw & 11108 & 29837 & 62131 \\
optima & confof & ekaw & 7424 & 22506 & 44528 \\
\hline
\end{tabular}
\end{center}
\caption{Ontologies diagnosable only with \textsc{Inv-HS-Tree}.}
\label{tab:unsolv30}
\end{table}

In the 13 cases presented in Table~\ref{tab:unsolv30} the \textsc{HS-Tree} was unable to find all requested diagnoses in each experiment. Within 2 hours the algorithm calculated only 1 diagnosis for \texttt{csa-conference-ekaw} and for \texttt{ldoa-conference-confof} it was able to find 1 and 9 diagnoses. The results of the \textsc{Inv-HS-Tree} are comparable with the presented in the Fig.~\ref{fig:solvable}. This experiment shows that the direct diagnosis is a stable and practically applicable method even in the cases when an ontology matching system outputs results of only moderate quality.

\begin{table}[tb]
\begin{center}
\begin{tabular}{|l|l|l|l|r|r|r|r|r|}
\hline
\textbf{Matcher} & \textbf{Ontology 1} & \textbf{Ontology 2} & \textbf{Scoring} & \textbf{Time (ms)} & \textbf{\#Query} & \textbf{React (ms)} & \textbf{\#CC} & \textbf{CC (ms)} \\
\hline
ldoa & conference & confof  & ENT & 11624 & 6 & 1473 & 430 & 3 \\
ldoa & conference & confof  & SPL & 11271 & 7 & 1551 & 365 & 4 \\
ldoa & cmt & ekaw  & ENT & 48581 & 21 & 2223 & 603 & 16 \\
ldoa & cmt & ekaw  & SPL & 139077 & 49 & 2778 & 609 & 54 \\
mappso & confof & ekaw  & ENT & 9987 & 5 & 1876 & 341 & 7 \\
mappso & confof & ekaw  & SPL & 31567 & 13 & 2338 & 392 & 21 \\
optima & conference & ekaw  & ENT & 16763 & 5 & 2553 & 553 & 8 \\
optima & conference & ekaw  & SPL & 16055 & 8 & 1900 & 343 & 12 \\
optima & confof & ekaw  & ENT & 23958 & 20 & 1137 & 313 & 14 \\
optima & confof & ekaw  & SPL & 17551 & 10 & 1698 & 501 & 6 \\
ldoa & conference & ekaw  & ENT & 56699 & 35 & 1458 & 253 & 53 \\
ldoa & conference & ekaw  & SPL & 25532 & 9 & 2742 & 411 & 16 \\
csa & conference & ekaw  & ENT & 6749 & 2 & 2794 & 499 & 3 \\
csa & conference & ekaw  & SPL & 22718 & 8 & 2674 & 345 & 20 \\
mappso & conference & ekaw  & ENT & 27451 & 13 & 1859 & 274 & 28 \\
mappso & conference & ekaw  & SPL & 70986 & 16 & 4152 & 519 & 41 \\
ldoa & cmt & edas  & ENT & 24742 & 22 & 1037 & 303 & 8 \\
ldoa & cmt & edas  & SPL & 11206 & 7 & 1366 & 455 & 2 \\
csa & conference & edas  & ENT & 18449 & 6 & 2736 & 419 & 5 \\
csa & conference & edas  & SPL & 240804 & 37 & 6277 & 859 & 36 \\
csa & edas & iasted  & ENT & 1744615 & 3 & 349247 & 1021 & 1333 \\
csa & edas & iasted  & SPL & 7751914 & 8 & 795497 & 577 & 11497 \\
ldoa & ekaw & iasted  & ENT & 23871492 & 10 & 1885975 & 287 & 72607 \\
ldoa & ekaw & iasted  & SPL & 20448978 & 9 & 2100123 & 517 & 37156 \\
mappso & edas & iasted  & ENT & 18400292 & 5 & 2028276 & 723 & 17844 \\
mappso & edas & iasted  & SPL & 159298994 & 11 & 13116596 & 698 & 213210 \\
\hline
\end{tabular}
\end{center}
\caption{Diagnosis discrimination using direct ontology debugging. \textbf{Scoring} stands for query selection strategy, \textbf{react} system reaction time between queries, \textbf{\#CC} number of consistency checks, \textbf{CC} gives average time needed for one consistency check.}
\label{tab:querysessionsinvhstree}
\end{table}

Moreover, in the first experiment we evaluated the efficiency of the interactive direct debugging approach applied to the cases listed in Table~\ref{tab:unsolv30}. In order to select the target diagnosis we searched for all possible minimal diagnoses of the following diagnosis problem instance $\tuple{M_{f}, \mo_i \cup \mo_j \cup M_t, \emptyset, \emptyset}$, where $M_{f}$ and $M_t$ are the sets of \emph{false} and \emph{true} positive alignments.  Both sets can be computed from the set of correct alignments $M_c$, provided by the organizers of OAEI 2011, and the set $M_{ij}$ generated by a ontology matching system as $M_{f} = M_{ij} \setminus M_c$ and $M_t = M_{ij} \cap M_c$. 
From this set of diagnoses we choose one diagnosis at random as the target. In the experiment the  prior fault probabilities of diagnoses were assumed to be $1-v$, where $v$ is the confidence value of the ontology matching system that the alignment is correct. Moreover, all axioms of both ontologies $\mo_i$ and $\mo_j$ were assumed to be correct and were assigned small probabilities.

The results presented in Table~\ref{tab:querysessionsinvhstree} were computed using both split-in-half (SPL) and entropy measure (ENT) for query selection for the diagnosis problem instance $\tuple{M_{ij}, \mo_i \cup \mo_j, \emptyset, \emptyset}$. The entropy measure was able to to solve the problem more efficiently because it is able to use information provided by the ontology matcher in terms on confidence values. The experiment shows also that efficiency any debugging methods depends highly on the ability of the underlying reasoner to classify an ontology. Note that the comparison of the suggested debugging technique with the ones build-in to such ontology matching systems as CODI~\cite{noessner2010} or LogMap~\cite{Jimenez-Ruiz2011} is inappropriate, since all these systems use greedy diagnosis techniques (e.g.~\cite{MeilickeStuck2009}), whereas the method presented in this paper is complete. However, the results presented in Table~\ref{tab:unsolv30} as well as in Fig.~\ref{fig:solvable} indicate that the suggested approach can find one minimal diagnosis in 25 seconds on average, which is comparable with the time of the mentioned systems.

In the second evaluation scenario we applied the direct method to unsatisfiable ontologies, generated for the Anatomy problem. The source ontologies $\mo_1$ and $\mo_2$ include 11545 and 4838 axioms correspondingly, whereas the size of the alignments varies between 1147 and 1461 axioms. The diagnosis selection process was performed in the same way as in the first experiment, i.e.\ we selected randomly one of the diagnoses of the instance $\tuple{M_{f}, \mo_1 \cup \mo_2 \cup M_t, \emptyset, \emptyset}$. The tests were performed for the problem instance diagnosis  $\tuple{M_{12}, \mo_1 \cup \mo_2, \emptyset, \emptyset}$ for 7 of the 12 systems. We excluded the results of CODI, CSA, MaasMtch, MapEVO and Aroma, because CODI produced coherent alignments and the output of the other systems was not classifiable within 2 hours. The results of the experiment show that the target diagnosis can be computed within 40 second in an average case. Moreover, \textsc{Inv-HS-Tree} slightly outperformed \textsc{HS-Tree}.

\begin{table}[tb]
\begin{center}
\begin{tabular}{|l|l|r|r|}
\hline
\textbf{Matcher} & \textbf{Scoring} & \textbf{\textsc{Inv-HS-Tree}} & \textbf{\textsc{HS-Tree}} \\
\hline
AgrMaker & ENT & 19.62 & 20.833 \\
AgrMaker & SPL & 36.035 & 36.034 \\
GOMMA-bk & ENT & 18.343 & 14.472 \\
GOMMA-bk & SPL & 18.946 & 19.512 \\
GOMMA-nobk & ENT & 18.261 & 14.255 \\
GOMMA-nobk & SPL & 18.738 & 19.473 \\
Lily & ENT & 78.537 & 82.524 \\
Lily & SPL & 82.944 & 115.242 \\
LogMap & ENT & 6.595 & 13.406 \\
LogMap & SPL & 6.607 & 15.133 \\
LogMapLt & ENT & 14.847 & 12.888 \\
LogMapLt & SPL & 15.589 & 17.45 \\
MapSSS & ENT & 81.064 & 56.169 \\
MapSSS & SPL & 88.316 & 77.585 \\
\hline
\end{tabular}
\end{center}
\caption{Scalability test for \textsc{Inv-HS-Tree}, time given in seconds.} 
\label{tab:scalability}
\end{table}

\section{Conclusions} \label{sec:conc}

In this paper we present an approach to direct computation of diagnoses for ontology debugging. By avoiding computation of conflict sets, the algorithms suggested in the paper are able to diagnose the  ontologies for which a common model-based diagnosis technique fails. Moreover, we show that the  method can also be used with diagnosis discrimination algorithms, thus, allowing interactive ontology debugging. The experimental results presented in the paper indicate that the performance of a system using the direct computation of diagnoses is either comparable with or outperforms the existing approach  in the settings, when faulty ontologies are generated by ontology matching or learning systems. The scalability of the algorithms was demonstrated on a set of big ontologies including thousands of axioms.

%\ack We would like to thank the referees for their comments, which helped improve this paper considerably
\bibliography{V-Know}
\end{document}